\documentclass[10pt,twocolumn,letterpaper]{article}

\usepackage{hyperref}
\usepackage{wacv}
\usepackage{times}
\usepackage{epsfig}
\usepackage{graphicx}
\usepackage{amsmath}
\usepackage{amssymb}
\usepackage{url}

\usepackage{tabularx,ragged2e}
\newcolumntype{C}{>{\Centering\arraybackslash}X}

\wacvfinalcopy 

\def\wacvPaperID{164} 

\ifwacvfinal\fi
\begin{document}

\title{ End-to-End Video Captioning with Multitask
Reinforcement Learning}

\author{Lijun Li\thanks{Work done in Tencent.}\\
Beihang University\\
{\tt\small lilijun1990@buaa.edu.cn}
\and
Boqing Gong\\
Tencent AI LAB\\
{\tt\small boqinggo@outlook.com}
}

\maketitle
\ifwacvfinal\thispagestyle{empty}\fi

\begin{abstract}
   Although end-to-end (E2E) learning has led to impressive progress on a variety of visual understanding tasks, it is often impeded by hardware constraints (e.g., GPU memory) and is prone to overfitting. When it comes to video captioning, one of the most challenging benchmark tasks in computer vision, those limitations of E2E learning are especially amplified by the fact that both the input videos and output captions are lengthy sequences. Indeed, state-of-the-art methods for video captioning process video frames by convolutional neural networks and generate captions by unrolling recurrent neural networks. If we connect them in an E2E manner, the resulting model is both memory-consuming and data-hungry, making it extremely hard to train. In this paper, we propose a multitask reinforcement learning approach to training an E2E video captioning model. The main idea is to mine and construct as many effective tasks (e.g., attributes, rewards, and the captions) as possible from the human captioned videos such that they can jointly regulate the search space of the E2E neural network, from which an E2E video captioning model can be found and generalized to the testing phase. To the best of our knowledge, this is the first video captioning model that is trained end-to-end from the raw video input to the caption output. Experimental results show that such a model outperforms existing ones to a large margin on two benchmark video captioning datasets.\footnote{Code available at \textcolor{red}{\url{https://github.com/adwardlee/multitask-end-to-end-video-captioning}}}
\end{abstract}

\section{Introduction}

Video captioning, i.e., to automatically describe videos by full sentences or phrases, not only serves as a challenging testbed in computer vision and machine learning but also benefits many real-world applications. The automatically generated video captions may enable fast video retrieval, assist the visually impaired, and engage users in a versatile chatbot, to name a few. 

Most recent works~\cite{venugopalan2015sequence,anne2016deep,hori2017attention,baraldi2016hierarchical} that tackle this problem fall under an encoder-decoder framework which has been shown effective in speech recognition~\cite{chan2016listen,toshniwal2017multitask}, natural language translation~\cite{johnson2016google,chung2016character}, and image captioning~\cite{xu2015show,liu2016optimization}. The encoder extracts compact representations of the visual content. In the context of video captioning, the convolutional neural networks (CNNs) are usually used to encode the video frames followed by a temporal model~\cite{hori2017attention,venugopalan2016improving,pan2016hierarchical,yao2016boosting} or simply temporal pooling~\cite{venugopalan2014translating} and the decoder maps the codes to a sequence of words often by the recurrent neural networks (RNNs)~\cite{rumelhart1985learning,werbos1988generalization} (e.g., the long short-term memory (LSTM)~\cite{hochreiter1997long} units are a popular choice). In order to train such networks, most existing works employ a cross-entropy loss at each decoding step. We refer the readers to the seminal work that spurs the resurging interests in video captioning, sequence to sequence - video to text (S2VT)~\cite{venugopalan2015sequence}, for a quick understanding about the backbone techniques. 

Despite the impact of the encoder-decoder framework on video captioning, it inherently impedes the use of end-to-end (E2E) training which has led to very impressive results on a large variety of tasks. Indeed, both CNNs and RNNs are memory consuming, leaving little GPU space to the training data which are yet key to the training procedure. Besides, the input videos and output sentences are both sequences, making the encoder-decoder framework very lengthy and data-hungry. On the one hand, it is tempting to explore the  E2E training strategy on the video captioning task. On the other hand, this seemingly straightforward idea is confined by the hardware and the relatively small size of existing video captioning datasets. Our experiments show that the conventional cross-entropy loss coupled with stochastic gradient descent cannot effectively exploit the E2E training.

\begin{figure*}[t]
    \centering
    \includegraphics[height=4in]{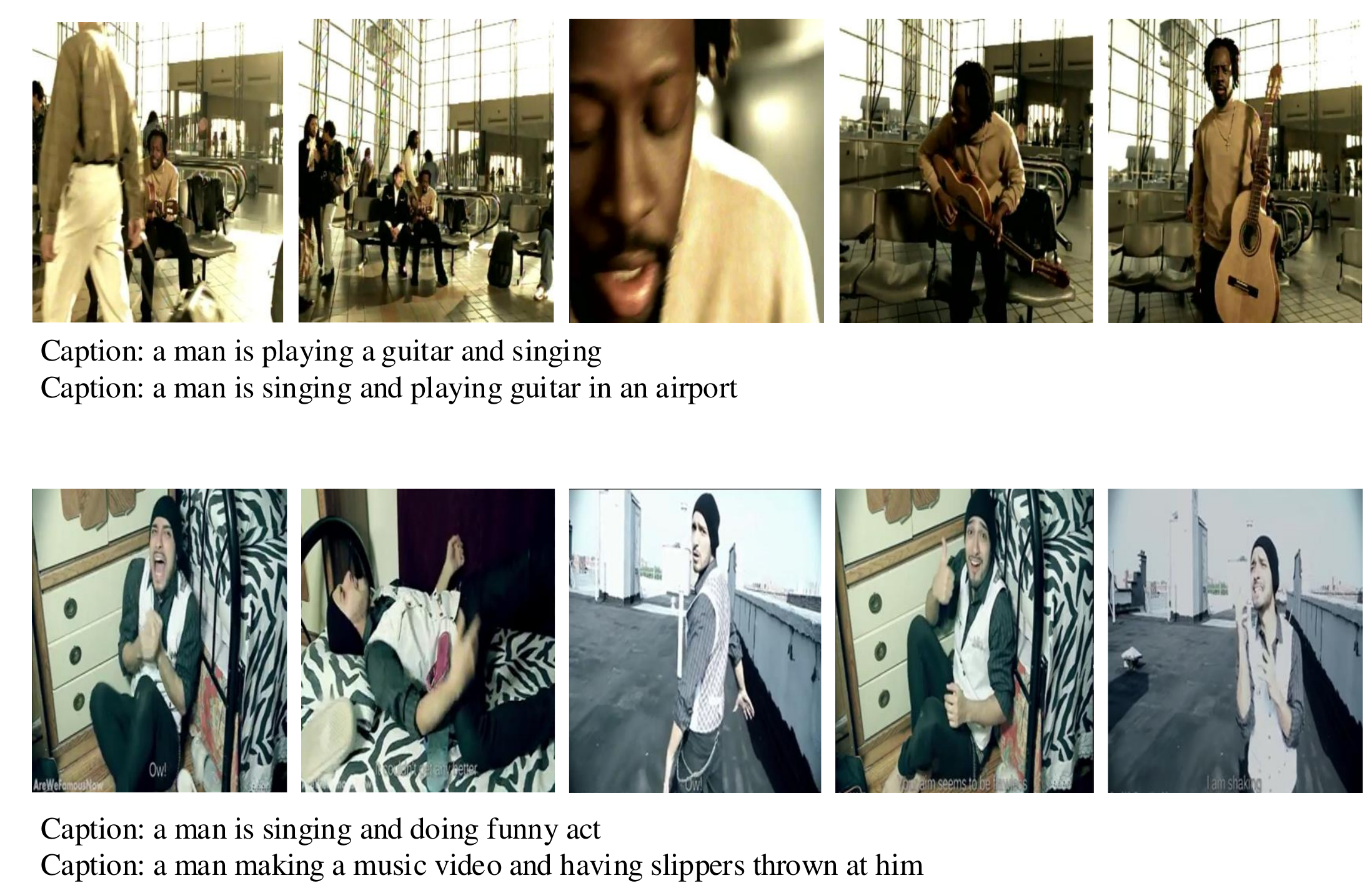}
    \caption{Exemplar video frames and user captions in the MSVD and MSR-VTT datasets.}
    \label{fig:examples}
\end{figure*}

In this paper, we propose a multitask reinforcement learning approach to training a video captioning model in an E2E manner. Our main idea is to mine and construct as many effective tasks as possible from the human captioned videos such that they can jointly regulate the search space of the encoder-decoder network, from which an E2E video captioning model can be found and generalized to the testing phase. The auxiliary tasks consist of two broad types: to predict the attributes extracted from the captions of the training videos and to maximize the rewards defined from the reinforcement learning perspective. When the training set is relatively small for the big encoder-decoder network, it is important to 
mine as much supervision as possible from the limited data so that it helps reduce the search space for the main task of interest.


Although many existing video captioning models~\cite{venugopalan2015sequence,pan2016video,yu2017end} can literally be trained in the E2E fashion, none of them were probably due to the hardware constraint and concerns on overfitting. Indeed, our study reveals that, without the proposed multitask reinforcement learning strategy, E2E learning is easy to overfit the training set. This work is the first to end-to-end train a model for video captioning, to the best of our knowledge. It is nontrivial because the model becomes very large in order to take as input a raw video sequence and output a sequence of words, causing challenges to the computational resources and raising the need for large-scale well-labeled data. Our multitask reinforcement learning method is able to alleviate those challenges and gives rise to state-of-the-art results on MSVD~\cite{chen2011collecting} and MSR-VTT~\cite{xu2016msr}, two popular benchmark datasets for the video captioning task. Nonetheless, we believe that, supplied with larger-scale labeled data, the E2E training  can further advance the video captioning results. 

We summarize our contribution as the following. (1) We propose a multitask reinforcement training strategy which can effectively learn a video captioning model in an E2E fashion under the current constraints of hardware and data size.  (2) We extract attributes from the captions of the training videos and define rewards upon the captions  without using any external data.  (3) Experiments show that our approach with a single model gives rise to state-of-the-art results on both the MSVD and MSR-VTT datasets.

\section{Related works}

Large amount of progress has been made in image and video captioning.  A large part of it is due to the advances in machine translation. For example, the encoder-decoder framework and the attention mechanism were first introduced in machine translation \cite{bahdanau2014neural,cho2014learning,sutskever2014sequence} and then extended to captioning. Both image captioning approaches~\cite{xu2015show,you2016image,chen2016sca} and video captioning methods~\cite{yao2015describing,hori2017attention,yu2016video,pu2016adaptive} follow their pipeline and also apply attention mechanism in caption generation. Comparing with image captioning, video captioning describes dynamic scenes instead of static scenes. From Figure~\ref{fig:examples}, we can clearly see that the video captioning is much more difficult with large variance in appearance. Baraldi et al. \cite{baraldi2016hierarchical} propose boundary-aware LSTM cell to automatically detect the temporal video segments. Venugopalan et al. \cite{venugopalan2016improving} integrate natural language knowledge to their network by training language LSTM model on a large external text corpora. Zhu et al. \cite{zhu2016bidirectional} extend Gated Recurrent Unit (GRU) to multirate GRU to handle different video frame rates. Hendricks et al. \cite{anne2016deep} propose a deep compositional captioner to describe novel object with the help of lexical classifier training on external image description dataset.

In the recent years, maximum likelihood estimation algorithm has been widely used in video captioning which maximizes the probability of current words based on the previous ground truth words \cite{gan2016semantic,donahue2015long,shen2017weakly,pan2016video,venugopalan2014translating}. But they all have two major problems.

The first one is exposure bias which is the input mismatch in training and inference. In training, the output of decoder depends on ground truth words instead of model predictions. While in inference, the decoder only has access to the predictions. Bengio et al. \cite{bengio2015scheduled} proposed scheduled sampling to mitigate the gap between the training and inference by selecting more often from the ground truth in the beginning but sampling more often from the model predictions in the end. However, it still optimizes at the word level.

The other problem is the objective mismatch between training and inference. In training, it optimizes the loss at the word level. While in inference, discrete metrics such as BLEU4 \cite{papineni2002bleu}, METEOR \cite{banerjee2005meteor}, CIDEr \cite{vedantam2015cider}, and ROUGE-L \cite{lin2004rouge} are used for evaluation. A few image captioning works have been proposed to solve the problems and shown superior performance with the help of reinforcement learning. Ren et al. \cite{ren2017deep} introduce actor-critic method to image captioning and also propose a new lookahead inference algorithm which has better performance than beam search. Liu et al. \cite{liu2016optimization} employ policy gradient method to optimize the SPIDEr score. Dai et al. \cite{dai2017towards} combine a conditional generative adversial network with policy gradient which can produce natural and diverse sentences. However, there are much less works using reinforcement learning in video captioning.

In this paper, we exploit the reinforcement learning in video captioning, especially for the jointly training of CNNs and LSTMs. Note that many video captioning models can actually be deployed in an end-to-end manner, such as~\cite{venugopalan2015sequence,pan2016video,yu2017end}, etc. Venugopalan et al. propose a stack of two LSTM networks ~\cite{venugopalan2015sequence}. Pan et al. propose a novel transfer unit to feed the semantic concept to LSTM~\cite{pan2016video}. Yu et al. develop a high-level word detector and semantic attention mechanism which combines the concept with caption decoder \cite{yu2017end}. However, they actually treat CNN as feature extractor and do not train the CNN part of their framework. On the contrary, our method trains the CNN and the other part together. 

Multitask learning is a kind of machine learning technique. During multitask learning, multiple tasks are solved at the same time with a shared representation and is especially useful with limited number of original data. It has been widely utilized not only in computer vision \cite{wang2009boosted,yuan2012visual,yan2016multi,gebru2017fine}, but also in natural language processing \cite{collobert2008unified}. It becomes a natural choice for us since the model capacity likely outweigh the existing datasets when we aim to update all its weights from the raw video input to the caption output. However, few works use multitask learning in video captioning. We explore the effectiveness and find the multitask learning can also be useful in video captioning.

\section{An E2E trained video captioning model}
We describe the end-to-end (E2E) trained video captioning model in this section. It is essentially a deepened version of the S2VT model~\cite{venugopalan2015sequence}. Despite its simplicity in concept, it is very challenging to train the whole big model to reach a good generalization capability onto the test sets. Both our experiments and an earlier attempt by Yu et al.~\cite{yu2017end} indicate that the gain of jointly training the CNNs and LSTMs is only marginal over fixing the CNNs as feature extractors, if we do not have an effective training approach. To this end, one important contribution of this paper is the batch of techniques presented below which we find useful when they are combined for training the E2E video captioning model.   

\begin{figure*}[t]
    \centering
    \includegraphics[scale=0.6]{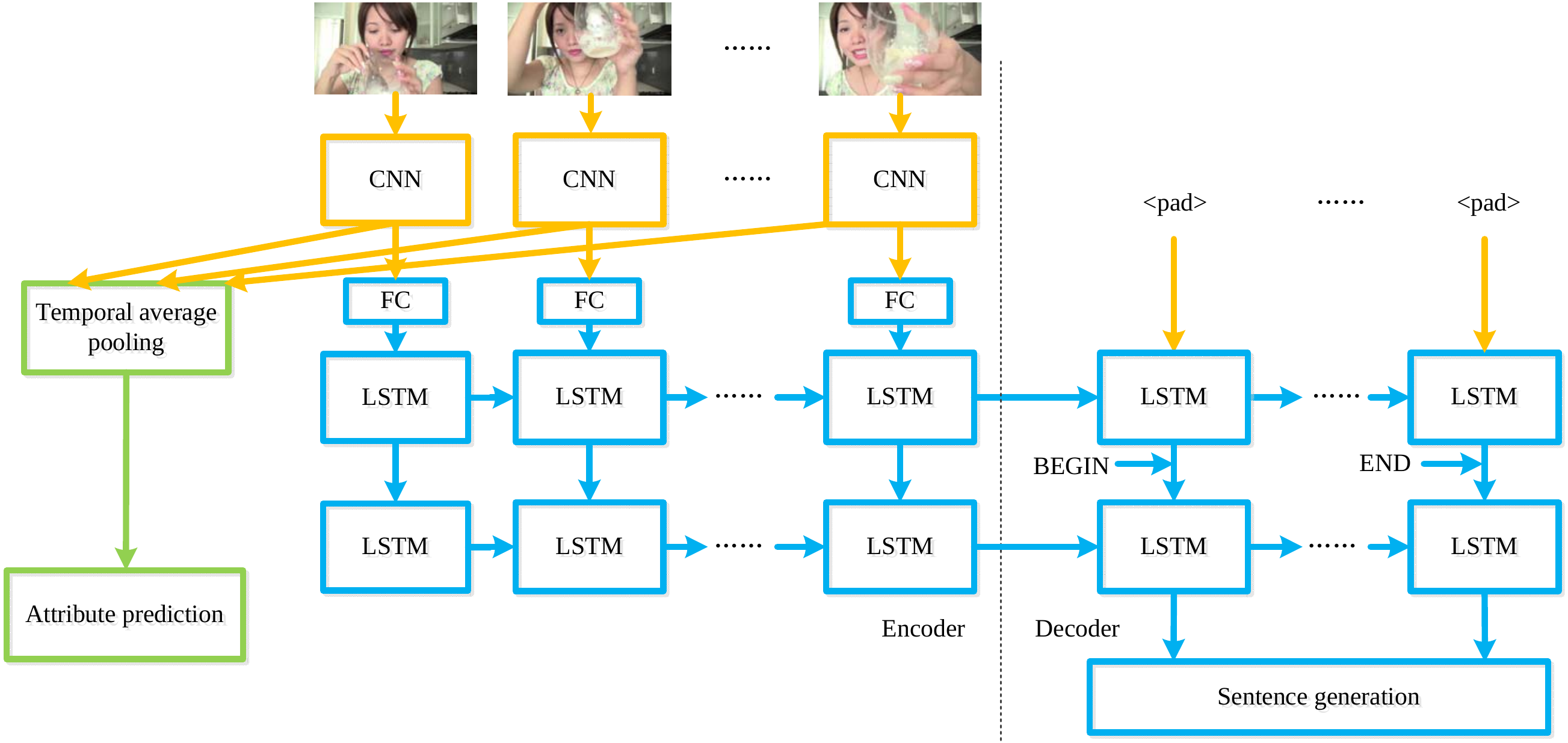}
    \caption{The multitask reinforcement learning framework for our E2E training of video captioning models. }
    \label{fig:multitask}
\end{figure*}

\subsection{Model architecture}
Figure~\ref{fig:multitask} sketches the model architecture which consists of three main components. On the top, five copies of the same Inception-Resnet-v2~\cite{szegedy2017inception} CNN are used to transform the raw video frames to high-level feature representations. Note that the last classification layer of the Inception-Resnet-v2 is replaced by a fully connected layer whose output dimension is 500. The LSTMs on the bottom first encode the video frames' feature representations and then decode a sentence to describe the content in the video. On the bottom left, there is a branch consisting of a temporal average pooling layer and an attribute prediction layer. We extract up to 400 attributes in our experiments. Accordingly, the attribute prediction layer's output dimension is 400 and the activation functions are sigmoid. This branch is introduced to assign relevant attributes to an input video. It is not used in the testing phase of the video captioning, but it generates informative gradients in the training phase for updating the weights of the CNNs in addition to those from the LSTMs. The design of the LSTMs (e.g., the number of hidden units, how to compute the input gates, etc.) is borrowed from S2VT~\cite{venugopalan2015sequence}.

\subsection{The E2E training of the model}
We train the model progressively in three steps. The first two steps aim to find a good initialization to the LSTMs (and the fully connected layer connecting the CNNs and the LSTMs) such that the last step, the E2E training of the whole model, can have a warm start. The weights of the CNNs are frozen until the third step.

\textbf{Step 1} is the standard training approach to S2VT using the cross-entropy loss. 
For an input frame $I_t$ at time step $t$, we encode it with the deep CNN and embed it with projection matrix $W_I$. Then for the projected feature representation $x_t$, the LSTM computes the hidden state $h_t$ and cell state $c_t$. The details about the computation of hidden state and cell state are in the following:
\begin{equation} \label{eq:1}
    \begin{aligned}
        i_t &= \sigma(W_{ix}x_t+W_{ih}h_{t-1}+b_i) \\
        f_t &= \sigma(W_{fx}x_t+W_{fh}h_{t-1}+b_f) \\
        o_t &= \sigma(W_{ox}x_t+W_{oh}h_{t-1}+b_o) \\
        g_t &= \phi(W_{gx}x_t+W_{gh}h_{t-1}+b_g) \\
        c_t &= i_t{\odot}g_t + f_t{\odot}c_{t-1} \\
        h_t &= o_t{\odot}{\phi}(c_t)
    \end{aligned}
\end{equation}
where ${\sigma}$ is the sigmoid function, ${\phi}$ is the hyperbolic tangent function, ${\odot}$ is element-wise multiplication. The second LSTM layer is similar to the first one, except that the input is the combination of first LSTM's output and the  word embeddings. 

Given a ``groundtruth'' sentence $s^\star=\{w_1^\star,w_2^\star,...w_T^\star\}$ describing an input video, we minimize the cross-entropy loss as follows,
\begin{equation}\label{eq:2}
    \begin{aligned}
    L_x(\theta):=-\log p_\theta(s^\star)=-\frac{1}{T}\sum_{t=1}^{T}\log p_\theta(w_t^\star|w_1^\star,\cdots,w_{t-1}^\star)
    \end{aligned}
\end{equation}
where $\theta$ denotes the model parameters. 

\paragraph{Step 2: REINFORCE+ training of the LSTMs.} After Step 1, we introduce the self-critical REINFORCE algorithm~\cite{williams1992simple,rennie2016self} to the video captioning to seek better weights for the LSTMs in terms of their generalization performance on the validation and test sets. 

It is well known that the cross-entropy loss exposes the recurrent LSTMs under different data distributions in the training and test stages because it feeds the model groundtruth words which are only available in training~\cite{bengio2015scheduled,rennie2016self}. Moreover, the loss function is not necessarily a good proxy for the evaluation metrics. To address these challenges, we opt to directly optimize the captioning system by REINFORCE learning as in \cite{rennie2016self}. In reinforcement learning, the goal is to train an agent to complete tasks by executing a series of actions in an environment. In the context of video captioning, the goal of the captioning model is to generate a proper sentence upon observing the video input. The captioning model corresponds to the agent and the action is to predict the next word at each time step. We can consider the input video with user annotated captions as the environment. We define the reward for the agent's action as the actual evaluation metric used in the test stage. In particular, we use the CIDEr score the as reward in this paper. Here is a brief summary of the reinforcement learning pipeline for video captioning: an agent receives an observation about the environment which contains the visual features and groundtruth words up to current step, as well as a reward (the CIDEr score) from the environment; the agent then takes an action to predict a word;  the environment provides another state (revealing one more groundtruth word) and reward in response to the agent's action.   

The objective function of reinforcement learning is:
\begin{equation}\label{eq:3}
    L_r(\theta)=-{\mathbb{E}}(r(w^s))
\end{equation}
where $w^s$ is the sentence consisting of $(w_1,w_2,...,w)$ sampled from the network and $r$ is the reward function. 

In order to solve the above problem, as in \cite{rennie2016self}, we also use the REINFORCE algorithm~\cite{williams1992simple}. The general updates of the parameter $\theta$ can be written as:
\begin{equation}\label{eq:4}
\triangledown_{\theta}L_r(\theta)=-{\mathbb{E}}[r(w)\triangledown{\log{p(w^s)}}],
\end{equation}
where $p(w^s)$ is basically determined by the video captioning model $p_\theta(w^s)$ (cf\ eq.~(\ref{eq:2})). In practice, the expectation is approximated by a sample mean which incurs variance to the gradients. To reduce the variance, the reward $r$ is often calibrated by a baseline $b$:
\begin{equation}\label{eq:5}
\triangledown_{\theta}L_r(\theta)=-{\mathbb{E}}[(r(w^s)-b)\triangledown_{\theta}\log{p_\theta{(w^s)}}],
\end{equation}
where it is obvious that the gradient remains unchanged since the baseline $b$ does not depend on the sampled words $w^s$. How to choose the baseline $b$ can affect the performance of the REINFORCE algorithm. We choose the reward of the greedily inferred words as our baseline. Denoting by $\widehat{w}_t:= \arg\max{p_\theta(w_t|h_t)}$, the baseline is $r(\widehat{w}^s)$. 

We are now ready to describe the practical algorithm for solving eq.~(\ref{eq:3}). A one-sample approximation to Eq. (\ref{eq:5}) is:
\begin{equation} \label{eq:6}
    \triangledown_{\theta}{L_r(\theta)}\approx{-(r(w^s)-b)\triangledown_{\theta}{\log{p_{\theta}(w^s)}}}
\end{equation}
which further be seen as the following cost function. At the beginning of each iteration, we sample up to $M$ trajectories (i.e., sentences) from the current model. Denoting them by $s_1,\cdots,s_M$, we can then write down the cost function for generating the gradients of this iteration, 
\begin{equation} \label{eq:8}
     {L_r(\theta)}\approx{-\frac{1}{M}\sum_{m=1}^{M}(r({s_m})-b){\log{p_{\theta}(s_m)}}}
\end{equation}
where $r(s_m)$ is the reward assigned to the trajectory $s_m$.  We denote this algorithm as REINFORCE+ or RFC+ in the following. 

It is interesting to note that Eq.~(\ref{eq:8}) acts as a running loss over the full course of the training. It changes at different iterations, being realized by the sampled trajectories as opposed to  the  constant  groundtruth  captions  in  the  cross-entropy loss ${L_x}$ across different iterations. Moreover, the rewards offset by the baseline dynamically weigh the contributions of the trajectories to the gradients. Jointly, they push the model trained in Step 1 further to the point that generalizes better to the unseen data.

\paragraph{Step 3: Multitask training of the full model.} We jointly tune the full model in this step, freeing the weights of the CNNs. As the starting point, it might seem natural to repeat Step 1 and/or Step 2 for the E2E optimization. However, this only gives rise to marginal gain over freezing the CNNs weights in our experiments. Such quick saturation of accuracy is actually common for very deep neural networks and may be alleviated by the skip connections between different layers of feedforward networks~\cite{he2016deep,srivastava2015highway}. Our model, however, heterogeneously mixes LSTMs and CNNs, leaving it unclear how to apply the skip connections. 

Instead, we propose to supply extra and informative gradients directly to the CNNs, so as to complement those reached to the CNNs indirectly through the LSTMs. Such direct gradients are provided by the attribute prediction branch (cf.\ Figure~\ref{fig:multitask}). 

We mine the attributes in the video captions following the previous practice on image captioning~\cite{wu2016value}. Among the words in the sentences of the training set, we extract the most frequent words  including nouns, verbs and adjectives as the attributes. Accordingly, the attribute prediction branch is equipped by sigmoid functions in order to each predict the existence or not ($y_i$) of an attribute in the input video. We define a binary cross entropy loss for this network branch, denoted by $L_a(\theta)=-\frac{1}{N}\sum_i \big[y_i\log q_\theta(i)+(1-y_i)\log (1-q_\theta(i))\big]$, where $N$ is the number of attributes in total and $q_\theta(i)$ is the network output for the $i$-th attribute. 

The overall cost function we use in Step 3 is a convex combination of the attribute loss and the REINFORCE loss:
\begin{equation}\label{eq:9}
    L(\theta) = {\alpha}L_r(\theta)+(1-\alpha)L_a(\theta)
\end{equation}
where $\alpha=0.95$ is selected by the validation set. 

\begin{table*}[ht] 
  \centering
  \caption{Comparison with state-of-the-art methods on the MSVD dataset.}
  \vspace{5pt}
  \begin{tabular}{l|c|c|c|c}
  \hline
  Models/Metrics & BLEU4 & ROUGE-L & METEOR & \textcolor{blue}{CIDEr} \\
  \hline
  \hline
        h-RNN \cite{yu2016video} & 0.499 & -- & 0.326 & \textcolor{blue}{0.658}  \\
        Attention fusion \cite{hori2017attention} & 0.524 & -- & 0.320 & \textcolor{blue}{0.688} \\
        BA encoder \cite{baraldi2016hierarchical} & 0.425 & -- & 0.324 & \textcolor{blue}{0.635} \\
        SCN \cite{gan2016semantic} & 0.502 & -- & \underline{\emph{0.334}} & \textcolor{blue}{0.770} \\
        TDDF \cite{zhangtask} & 0.458 & \underline{\emph{0.697}} & \underline{\emph{0.333}} & \textcolor{blue}{0.730}  \\
        LSTM-TSA \cite{pan2016video} & \underline{\emph{0.528}} & -- & \underline{\emph{0.335}} & \textcolor{blue}{0.740} \\
        MVRM \cite{zhu2016bidirectional} & \textbf{0.538} & -- & \textbf{0.344} & \textcolor{blue}{0.812} \\
    \hline    
    S2VT (our Step 1) \cite{venugopalan2015sequence} & 0.428 & 0.687 & 0.325 & \textcolor{blue}{0.750} \\
    REINFORCE (our Step 2) \cite{rennie2016self} & 0.456 & 0.690 & 0.329 & \textcolor{blue}{0.806} \\
    REINFORCE+ (our Step 2) \cite{rennie2016self} & 0.466 & \underline{\emph{0.694}} & 0.330 & \textcolor{blue}{0.816} \\
    \hline
    E2E (ours, greedy search) & 0.480 &  \textbf{0.705} & \underline{\emph{0.336}}  &  \textcolor{blue}{\underline{\emph{0.865}}}\\
     E2E (ours, beam search) & 0.503 & \textbf{0.708} & \textbf{0.341} & \textcolor{blue}{\textbf{0.875}}\\
  \hline
  \end{tabular}
  \label{tab:results_msvd}
\end{table*}

\begin{table*}[ht]
    \centering
    \caption{Comparison with state-of-the-art methods on the MSR-VTT dataset.}
    \vspace{5pt}
    \begin{tabular}{l|c|c|c|c}
    \hline
     Models & BLEU4 & ROUGE-L & METEOR & \textcolor{blue}{CIDEr} \\
    \hline
    \hline
        TDDF \cite{zhangtask} & 0.372 & 0.586 & \underline{\emph{0.277}} & \textcolor{blue}{0.441}  \\
        v2t\_navigator\cite{jin2016describing} & \textbf{0.408} & \underline{\emph{0.609}} & \textbf{0.282} & \textcolor{blue}{0.448}  \\
        Aalto \cite{shetty2016frame} & \underline{\emph{0.398}} & 0.598 &  0.269 & \textcolor{blue}{0.457} \\
        Attention fusion \cite{hori2017attention} & \underline{\emph{0.394}} & -- & 0.257 & \textcolor{blue}{0.404} \\
        \hline
    S2VT (our Step 1) \cite{venugopalan2015sequence} & 0.353 & 0.578 & 0.266 & \textcolor{blue}{0.407} \\
    REINFORCE (our Step 2) \cite{rennie2016self} & 0.392 & 0.603 & 0.267 & \textcolor{blue}{0.448} \\
    REINFORCE+ (our Step 2) \cite{rennie2016self}& \underline{\emph{0.398}} & \underline{\emph{0.609}} & 0.271 & \textcolor{blue}{\underline{\emph{0.468}}} \\
    \hline
    E2E (ours, greedy search) & \textbf{0.404} &  \textbf{0.610} & 0.270  & \textcolor{blue}{\textbf{0.483}}\\
    E2E (ours, beam search) & \textbf{0.404} & \textbf{0.610} & 0.270 & \textcolor{blue}{\textbf{0.483}}\\
    \hline
    \end{tabular}
    \label{tab:results_msrvtt}
    \vspace{0pt}
\end{table*}

\begin{table}
  \centering
  \caption{Ablation study: video captioning results on MSVD with greedy decoding.}
  \vspace{5pt}
  \resizebox{0.5\textwidth}{!}{
  \begin{tabular}{l|c|c|c|c}
  \hline
  Models & BLEU4 & ROUGE-L & METEOR & \textcolor{blue}{CIDEr} \\
  \hline
  \hline
    S2VT (Step 1) \cite{venugopalan2015sequence} & 0.428 & 0.687 & 0.325 & \textcolor{blue}{0.750} \\
    RFC (Step 2) \cite{rennie2016self} & 0.456 & \underline{\emph{0.690}} & \underline{\emph{0.329}} & \textcolor{blue}{0.806} \\
    RFC+ (Step 2) \cite{rennie2016self} & \underline{\emph{0.466}} & \underline{\emph{0.694}} & \underline{\emph{0.330}} & \textcolor{blue}{\underline{\emph{0.816}}} \\
    E2E (xentropy) & 0.439 & \underline{\emph{0.690}} & \underline{\emph{0.328}} & \textcolor{blue}{0.767} \\
    E2E (att+xentropy) & 0.453 & \underline{\emph{0.694}} & \underline{\emph{0.331}} & \textcolor{blue}{0.790}  \\
    E2E w/o attribute prediction & \underline{\emph{0.466}} & 0.696 & \underline{\emph{0.332}} & \textcolor{blue}{\underline{\emph{0.824}}}\\
    E2E w/o reinforcement or attribute or Step 1 & 0.424 & 0.684 & 0.325 & \textcolor{blue}{0.719} \\
    \hline
    E2E (ours) & \textbf{0.480} & \textbf{0.705} & \textbf{0.336} & \textcolor{blue}{\textbf{0.865}} \\
  \hline
  \end{tabular}}
  \label{tab:my_msvd}
\end{table}

\begin{table}
    \centering
    \caption{Ablation study: video captioning results on MSR-VTT dataset with greedy decoding.}
    \vspace{5pt}
    \resizebox{0.5\textwidth}{!}{
    \begin{tabular}{l|c|c|c|c}
    \hline
     Models & BLEU4 & ROUGE-L & METEOR & \textcolor{blue}{CIDEr} \\
    \hline
    \hline
    S2VT (Step 1) \cite{venugopalan2015sequence} & 0.353 & 0.578 & \underline{\emph{0.266}} &\textcolor{blue}{0.407} \\
    RFC (Step 2) \cite{rennie2016self} & 0.392 & \underline{\emph{0.603}} & \textbf{0.267} & \textcolor{blue}{0.448} \\
    RFC+ (Step 2) \cite{rennie2016self}& \underline{\emph{0.398}} & \textbf{0.609} & \textbf{0.271} & \textcolor{blue}{\underline{\emph{0.468}}} \\
    \hline
    E2E (ours) & \textbf{0.404} & \textbf{0.610} & \textbf{0.270} & \textcolor{blue}{\textbf{0.483}} \\
    \hline
    \end{tabular}}
    \label{tab:my_msrvtt}
    \vspace{-5pt}
\end{table}


\begin{table*}[ht]
    \centering
    \caption{Qualitative results of video captioning on MSVD dataset. Baseline is the sentence generated by our baseline model, MR stands for sentence generated by our multitask reinforce model and GT represents Ground Truth captions}
    \begin{tabularx}{\textwidth}{XXX}
    \begin{minipage}{.32\textwidth}
      \includegraphics[width=\linewidth, height=20mm]{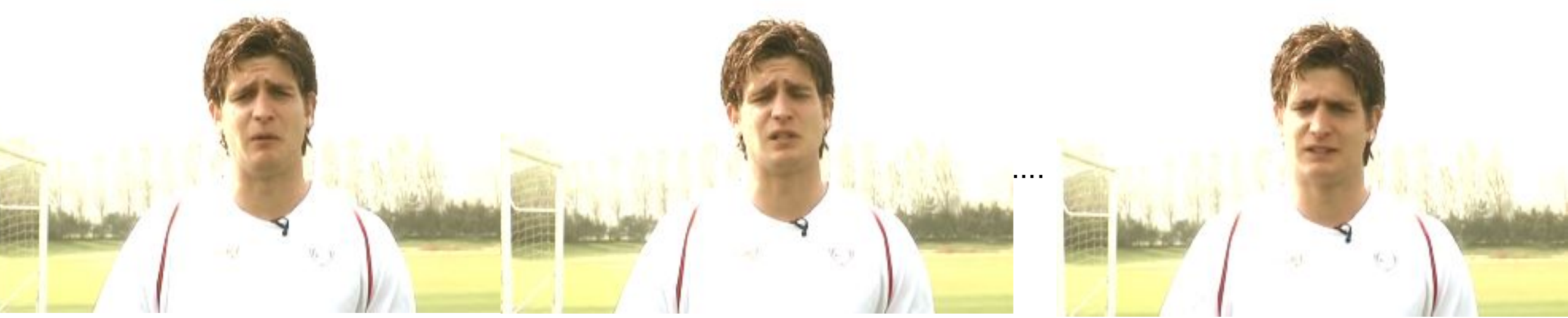}
    \end{minipage}
    
    &
    \begin{minipage}{.32\textwidth}
      \includegraphics[width=\linewidth, height=20mm]{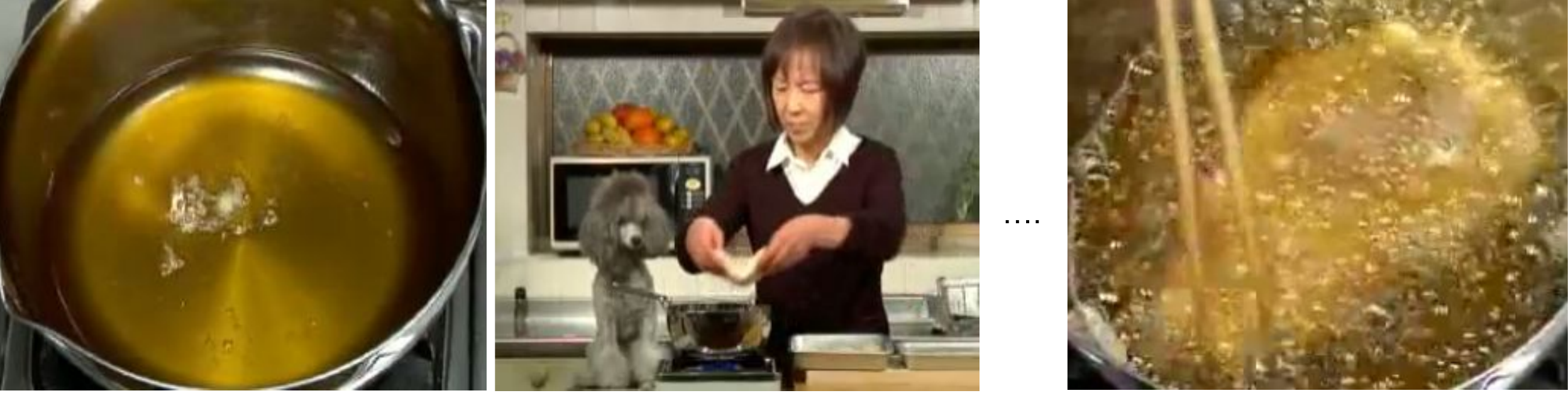}
    \end{minipage}
    &
    \begin{minipage}{.32\textwidth}
      \includegraphics[width=\linewidth, height=20mm]{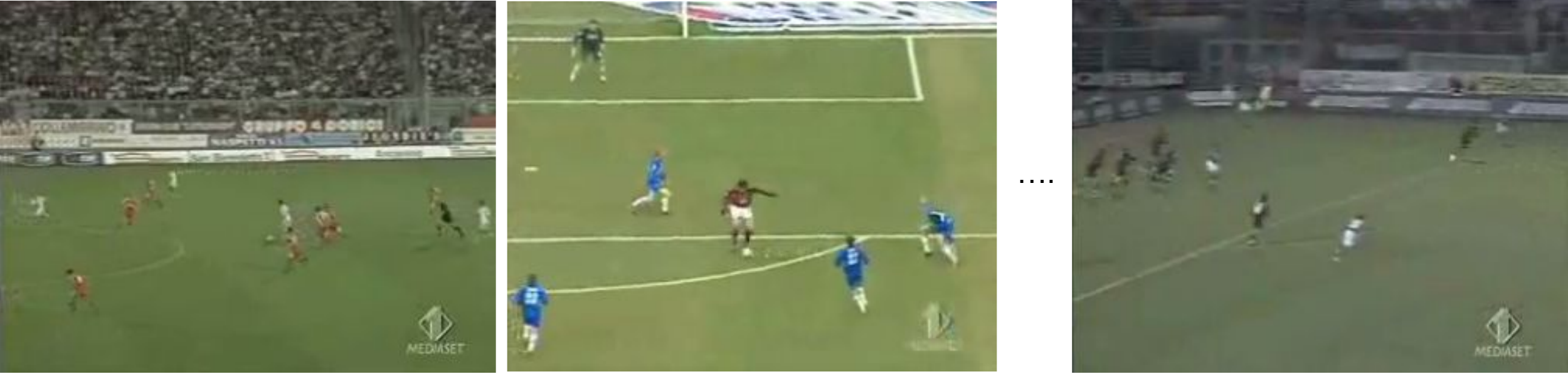}
    \end{minipage}\\
         \footnotesize{Captions: }&  \footnotesize{Captions: }& \footnotesize{Captions: }\\
           \footnotesize{S2VT: a man is giving a woman}& \footnotesize{S2VT: a woman is putting some meat in a pan} & \footnotesize{S2VT: a soccer player is kicking a soccer ball}\\
         \footnotesize{ E2E: a man is talk }& \footnotesize{E2E: a woman is frying  meat }& \footnotesize{E2E: men are playing soccer}\\
       \footnotesize{GT: a man is talking }& \footnotesize{GT: a woman is frying meat }& \footnotesize{GT: the men are playing soccer} \\
    \end{tabularx}
    \label{tab:caption_msvd}
\end{table*}

\begin{table*}[ht]
    \centering
    \caption{Qualitative results of video captioning on MSR-VTT dataset. Baseline is the sentence generated by our baseline model, MR stands for sentence generated by our multitask reinforce model and GT represents Ground Truth captions}
    \begin{tabularx}{\textwidth}{XXX}
    \begin{minipage}{.32\textwidth}
      \includegraphics[width=\linewidth, height=20mm]{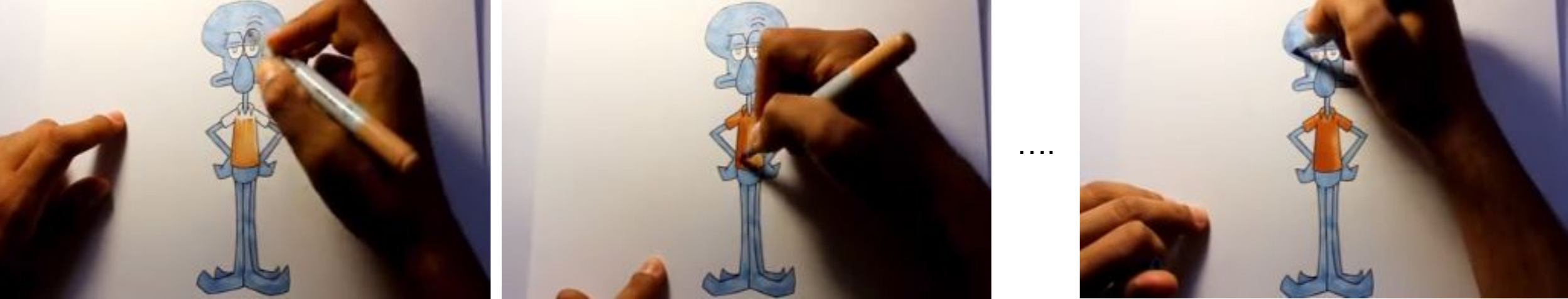}
    \end{minipage}
    
    &
    \begin{minipage}{.32\textwidth}
      \includegraphics[width=\linewidth, height=20mm]{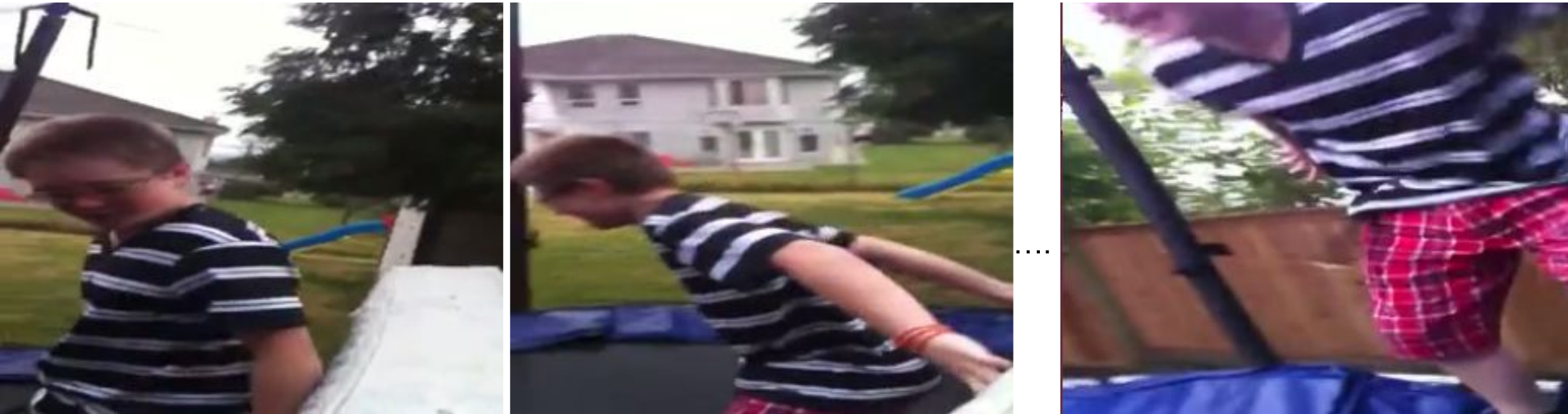}
    \end{minipage}
    &
    \begin{minipage}{.32\textwidth}
      \includegraphics[width=\linewidth, height=20mm]{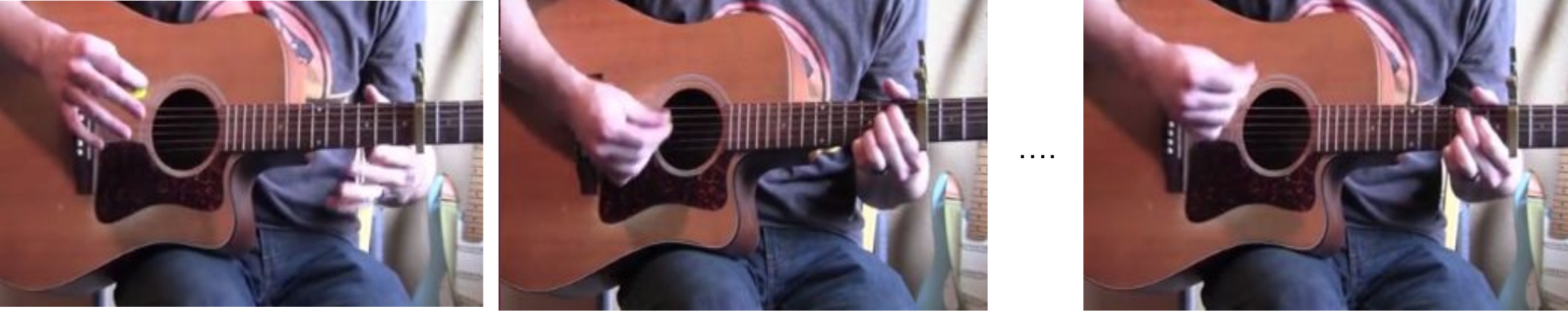}
    \end{minipage}
    \\
         \footnotesize{Captions: } &  \footnotesize{Captions: }& \footnotesize{Captions: }\\
         \footnotesize{S2VT: a person is talking about a computer}&  \footnotesize{S2VT: a man is singing } & \footnotesize{S2VT: a man is talking about a guitar}\\
         \footnotesize{E2E: a person is drawing a cartoon }& \footnotesize{E2E: a man is jumping on a trampoline }& \footnotesize{E2E: a man is playing a guitar}\\
         \footnotesize{GT: a person is drawing a cartoon } & \footnotesize{GT: a man is jumping on a trampoline }& \footnotesize{GT: a man is playing a guitar }\\
    \end{tabularx}
    \label{tab:caption_msrvtt}
\end{table*}

\section{Comparison experiments}
We present extensive experimental results and ablation studies in this section. 

\subsection{Datasets and experiment details} 
In this section, we report the results of our E2E trained model and compare with other state-of-the-art methods on two popular video captioning datasets. One is the MSVD dataset~\cite{chen2011collecting}. MSVD consists of 1,970 video clips and 70,028 captions collected via Amazon Mechanical Turk (\url{www.mturk.com}) which covers a lot of topics. On average, the video duration is about 10 seconds and each sentence contains about 8 words. A common split of the videos is provided by~\cite{venugopalan2015sequence} and maintained by the existing works as well as this paper: 1,200 videos for training, 100 for validation, and 670 for testing. The other is the MSR-VTT dataset which contains 10,000 video clips and 200,000 captions. We use the data split defined in~\cite{xu2016msr} in our experiments: 6,513 videos for training, 497 for validation, and 2,990 for testing. It is the largest publicly available video captioning dataset in terms of the number of sentences. The average duration of the videos is 20 seconds. 

\paragraph{Implementation details.} We implement our algorithm with Tensorflow~\cite{abadi2016tensorflow}. In our end-to-end trained model, we keep the layers of Inception-Resnet-v2~\cite{szegedy2017inception} until the last pooling layer whose dimension is 1,536. After that, we add a fully connected layer whose output dimension is 500.  The dimension of the LSTM hidden layers is 1000. A dropout layer is attached to each LSTM unit during training with dropout rate of 0.2. Each word is represented as one-hot vector. The image embedding dimension and word embedding dimension are both 500. We fix the encoder step size to 5 and decoder step size to 35. All the trainable parameters are initialized by drawing from the uniform distribution $[-0.1,0.1]$. The ADAM optimizer is used in our experiments. The learning rate is 1e-4 to train S2VT. For other methods, it is 1e-6. The hyperparameter $\alpha$ is 0.95 in Eq.~(\ref{eq:9}). For both datasets, we resize the video frames to 224x224. For inference, we use beam search to keep multiple generated words at current time step and select the best sentence with the beam size 3 in the end. All the free parameters are chosen by the validation sets. For the evaluation metrics, we choose four types of widely used caption metrics: BLEU4 \cite{papineni2002bleu}, METEOR \cite{banerjee2005meteor}, CIDEr \cite{vedantam2015cider}, and ROUGE-L \cite{lin2004rouge}. The scores are calculated using the MS COCO evaluation code~\cite{chen2015microsoft}.

\subsection{Baseline methods} 
Table~\ref{tab:results_msvd} and~\ref{tab:results_msrvtt} show the comparison results with several recently proposed methods on the two datasets, respectively. On the MSVD dataset, we compare our approach with following seven recent methods.

\textbf{h-RNN}~\cite{yu2016video} proposes a hierarchical-RNN framework and designs an attention scheme over both temporal and spatial dimensions to focus on the visual elements.

\textbf{Attention fusion}~\cite{hori2017attention} develops a modality-dependent attention mechanism together with temporal attention to combines the cues of multiple modalities, which can attend not only  time but also the modalities.

\textbf{BA encoder}~\cite{baraldi2016hierarchical} presents a new boundary-aware LSTM cell to detect the discontinuity of consecutive frames. Then the cell is used to build a hierarchical encoder and makes its structure adapt to the inputs. 

\textbf{SCN}~\cite{gan2016semantic} detects semantic concepts from videos and proposes a tag-dependent LSTM whose weights matrix depends on the semantic concepts. 

\textbf{TDDF}~\cite{zhangtask} combines motion feature and appearence feature, and automatically determines which feature should be focused according to the word. 

\textbf{LSTM-TSA}~\cite{pan2016video} presents a transfer unit to control and fuse the attribute, motion, and visual features for the video representations.

\textbf{MVRM}~\cite{zhu2016bidirectional} learns a multirate video representation which can adaptively fit the  motion speeds in videos. 

On the MSR-VTT dataset, we include four methods in the comparison: TDDF~\cite{zhangtask}, v2t\_navigator \cite{jin2016describing}, Aalto~\cite{shetty2016frame}, and Attention fusion~\cite{hori2017attention}. 

\textbf{V2t\_navigator}~\cite{jin2016describing} represents the videos by their visual, aural, speech, and category cues, while we only employ the raw video frames in our approach.

\textbf{Aalto}~\cite{shetty2016frame} trains an evaluator network to drive the captioning model towards semantically interesting sentences. 

During the experiments, REINFORCE (RFC) denotes the vailla self-critical REINFORCE algorithm extended to video captioning, and REINFORCE+ (RFC+) represents our  REINFORCE algorithm with multi-sampling trajectories. We denote by E2E our final multitask reinforcement learning approach (cf.\ Eq.~(\ref{eq:9})).

\subsection{Comparison results} Table~\ref{tab:results_msvd} and~\ref{tab:results_msrvtt} present the results evaluated by BLEU4 \cite{papineni2002bleu}, METEOR \cite{banerjee2005meteor}, CIDEr \cite{vedantam2015cider}, and ROUGE-L \cite{lin2004rouge} on MSVD and MSR-VTT, respectively. The CIDEr scores on MSVD dataset are much higher than those on MSR-VTT dataset. The reason may due to the much complex scenes, actions, and large variance in the MSR-VTT dataset. Our approach is denoted by E2E and two decoding results are reported, one by the greedy search and the other by the beam search of a window size of three. We can see that our approach outperforms the existing ones to  large margins under the CIDEr metric, which is taken as the reward function in our training procedure. Under the other metrics, ours is also among the top performing methods while we notice that one can conveniently replace the CIDEr reward by the other metrics as the reward functions. On MSVD dataset, our E2E method can acheive 0.865 in CIDEr. Comparing with the baseline method, S2VT, our E2E model can make a relative improvement by 15.3\% in CIDEr with greedy decoding. On the MSR-VTT dataset, our E2E method can reach 0.483 in CIDEr. It can make a relative improvement by 18.6\%.

Several factors may have contributed to the superior performance of our model. First, we fine-tune the CNNs such that the extracted features of the video frames are purposely tailored for the video captioning task, as opposed to the generic features pre-trained from the ImageNet. Besides, the LSTM architecture inherently exploits the temporal nature of the videos. At last but not the least, the performance attributes to the progressive and multitask techniques of  training the model. Next, we provide in-depth analyses about the last point by some ablation studies.

\subsection{Ablation study}
Due to the time and computation resource constraint, we mainly run the ablation experiments on the MSVD dataset. See Table~\ref{tab:my_msvd} for the results. Additionally, we also report some key results on MSR-VTT in Table~\ref{tab:my_msrvtt}. 

First of all, we note that Step 2 is able to significantly improve the results of Step 1, reinforcing the effectiveness of the REINFORCE algorithm~\cite{williams1992simple}. Besides, by sampling multiple trajectories (cf.\ row RFC+) we can boost the original REINFORCE by 1\% to 2\%.  

If we skip Step 2 and directly fine-tune the CNNs using the cross-entropy loss $L_x$ in Step 3, the results (cf.\ row E2E (xentropy)) are only marginal better than those of freezing the CNNs. This observation is not surprising, given that the full model is actually both deep in CNNs and long in terms of the unrolled LSTMs, making it very hard to train. 

If we skip Step 2 and instead minimize the convex combination of the attribute prediction loss $L_a$ and the cross entropy loss $L_x$ of the video captions, the results are much better than those of Step 1 and yet still worse than Step 2's. Hence, we conclude that 1) the attribute prediction branch helps the video captioning task and 2) the REINFORCE training is inevitable for eliminating the exposure mismatch~\cite{bengio2015scheduled} of the LSTMs between the training and testing stages.

If we remove the attribute prediction branch from our model and only use the REINFORCE+ to fine-tune the CNNs in Step 3, the results can only be very slightly improved which can be seen from E2E w/o attribute. This verifies the necessity of the attribute prediction branch. Indeed, this branch back-propagates the gradients from an albeit different attribute prediction task directly to the CNNs, being able to complement the gradients coming through the LSTM branch. If we do not follow the steps and directly train the model end-to-end, the result is even lower than S2VT. It confirms our E2E progressive training pipeline is effective.

\section{Qualitative results}
In Table \ref{tab:caption_msvd} and Table \ref{tab:caption_msrvtt}, a few video caption instances are shown of the MSVD dataset and the MSR-VTT dataset. The captions are generated by the S2VT model and our E2E model, respectively. We compare the sentences with the ground truth sentences in the Tables. Generally, our E2E model can generate relevant sentences. The sentences generated by our E2E model can reflect the visual content more faithfully with less grammar errors. For instance, our multitask model generates ``a woman is frying meat'' and it shows exactly what the woman is doing in the middle image of Table \ref{tab:caption_msvd}. It is more reasonable and relevant to the video content than ``a woman is putting some meat in a pan'' generated by the baseline model. Our E2E model also describe the event correctly, it recognizes there are a group of players playing soccer instead of one player in the right image of Table \ref{tab:caption_msvd}. On the other dataset, the examples also illustrate the correctness and faithfulness of our method. It can correctly detect drawing a cartoon compared to talking about a computer, the action of jumping on a trampoline and playing instead of talking about guitar. Under most test cases, our model is descriptive and more accurate than the baseline.


\section{Conclusion}
We propose a novel method which combines the reinforcement learning with attribute prediction to train the whole framework end-to-end for video captioning. For our E2E model, it is a multitask end-to-end network and combines multisampling reinforce algorithm to generate captions. It is the first time that the CNNs are learned together with RNNs in video captioning and show much improvement, to best of our knowledge. The experiments on two standard video captioning datasets show our model can outperform the current methods. It also shows that the domain adopted video representation is more powerful than the generic image features. In the future, we will explore more representative video representations. As the 3D convolution methods are effective in the video classification, e.g I3D~\cite{carreira2017quo}, we believe our model can also benefit from employing the effective video representation in video classification field. We may also explore other multitasks to better fine-tune the video representation.


{\small
\bibliographystyle{ieee}
\bibliography{egbib}
}

\end{document}